\documentclass[10pt,twocolumn,letterpaper]{article}

\usepackage[pagenumbers]{cvpr} 

\usepackage{graphicx}
\usepackage{amsmath}
\usepackage{amssymb}
\usepackage{booktabs}

\usepackage{xcolor}
\usepackage{multirow}
\usepackage{stfloats}
\usepackage{adjustbox}

\newcommand\modelname[1]{HPL-ESS}

\definecolor{cvprblue}{rgb}{0.21,0.49,0.74}
\usepackage[pagebackref,breaklinks,colorlinks,citecolor=cvprblue]{hyperref}
\usepackage[capitalize]{cleveref}
\def\paperID{12305} 
\def\confName{CVPR}
\def\confYear{2024}

\begin{document}

\title{HPL-ESS: Hybrid Pseudo-Labeling for Unsupervised \\ Event-based Semantic Segmentation }

\author{Linglin Jing$^{1,2}$\thanks{Equal contribution.}, Yiming Ding$^{1*}$, Yunpeng Gao$^{1,4}$, Zhigang Wang$^{1}$\thanks{Corresponding author.}, Xu Yan$^{3}$,\\Dong Wang$^{1}$, Gerald Schaefer$^{2}$,Hui Fang$^{2\dagger}$, Bin Zhao$^{1,4}$, Xuelong Li$^{1,5}$\\
$^{1}$Shanghai AI Laboratory,
$^{2}$Loughborough University,
$^{3}$SSE \& FNII, CUHK-Shenzhen,\\
$^{4}$Northwestern Polytechnical University,
$^{5}$Institute of Artificial Intelligence (TeleAI)
\\
{\tt\small l.jing@lboro.ac.uk, wangzhigang@pjlab.org.cn}
}
\maketitle
\begin{abstract}

Event-based semantic segmentation has gained popularity due to its capability to deal with scenarios under high-speed motion and extreme lighting conditions, which cannot be addressed by conventional RGB cameras. Since it is hard to annotate event data, previous approaches rely on event-to-image reconstruction to obtain pseudo labels for training. However, this will inevitably introduce noise, and learning from noisy pseudo labels, especially when generated from a single source, may reinforce the errors. This drawback is also called confirmation bias in pseudo-labeling. 
In this paper, we propose a novel hybrid pseudo-labeling framework for unsupervised event-based semantic segmentation, \modelname{}, to alleviate the influence of noisy pseudo labels. Specifically, we first employ a plain unsupervised domain adaptation framework as our baseline, which can generate a set of pseudo labels through self-training. Then, we incorporate offline event-to-image reconstruction into the framework, and obtain another set of pseudo labels by predicting segmentation maps on the reconstructed images. A noisy label learning strategy is designed to mix the two sets of pseudo labels and enhance the quality. Moreover, we propose a soft prototypical alignment (SPA) module to further improve the consistency of target domain features. Extensive experiments show that the proposed method outperforms existing state-of-the-art methods by a large margin on benchmarks (\emph{e.g.}, +5.88\% accuracy, +10.32\% mIoU on DSEC-Semantic dataset), and even surpasses several supervised methods. 

\end{abstract}

\section{Introduction}
\label{sec:intro}

\begin{figure}[!t] 
\begin{center}
\includegraphics[width=0.95\linewidth]{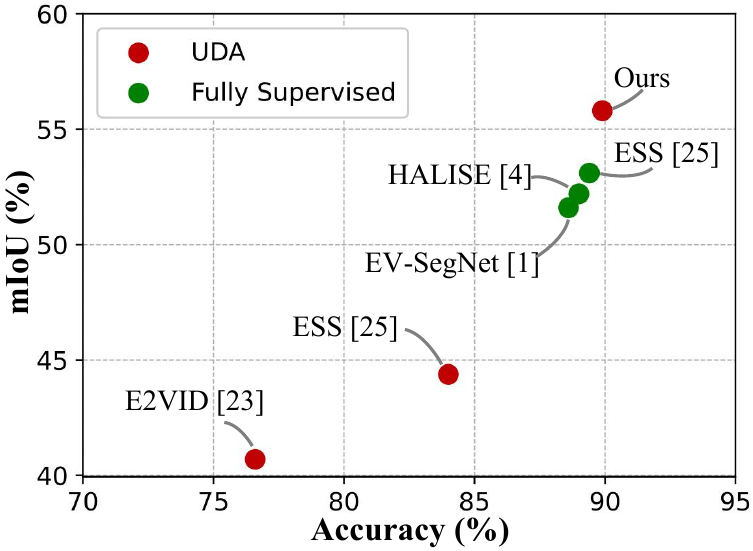}
\end{center}
\caption{Comparison on the DSEC-Semantic dataset. Our method outperforms other UDA works by a large margin and even surpasses fully supervised methods.}
\label{fig:preview}
\end{figure}

Event cameras are bio-inspired vision sensors that respond to changes in pixel intensity, generating a stream of asynchronous events characterized by exceptionally high temporal resolution. This technology enables the capture of dynamic scenes, providing features of high dynamic range (HDR) and reduced motion blur.
Event cameras have been extensively applied in various applications, including object recognition~\cite{orchard2015hfirst,jing2022towards}, SLAM~\cite{gallego2017event}, and autonomous driving systems~\cite{maqueda2018event}, effectively addressing challenges such as motion blur and overexposure.

However, event data significantly differ from images, making it difficult to annotate in dense pixel prediction tasks such as semantic segmentation.
Previous works~\cite{wang2021evdistill, alonso2019ev, wang2021dual} require per-pixel paired events and images, and then leverage pre-trained networks on images to generate labels for event data.
Although a more precisely paired and sharper image would naturally yield improved results, these methods increase the demands on capture devices.
Other methods rely on event-to-image conversion to get rid of the need for ground-truth labels.  E2VID~\cite{rebecq2019high} is an event-to-image (ETI) reconstruction method to transform events into images, while VID2E~\cite{gehrig2020video} employs image-to-event (ITE) in reverse.
Based on the above methods, a feasible strategy is to generate pseudo labels from converted images for event data. ESS~\cite{sun2022ess} further employs unsupervised domain adaptation (UDA) to transfer knowledge from labeled image data (source domain) to unlabeled event data (target domain) through the bridge of event-to-image reconstruction.

Despite improvements, reconstruction-based methods suffer from the limitation that, due to the lack of texture information in event data, the reconstructed images usually have large fuzzy regions, inevitably introducing noise into the generated pseudo labels. 
Training on noisy pseudo labels has the risk of reinforcing the errors, especially when they are obtained from a single source, a problem that is known as confirmation bias~\cite{ArazoOAOM20} in pseudo-labeling.

To alleviate the bias of single-source pseudo labels, in this paper, we propose \modelname{}, a hybrid pseudo-labeling framework for unsupervised event-based semantic segmentation. Our method is built upon a modified UDA framework, which executes self-training on the mixture of unpaired images and event data. The framework has the ability to generate a set of pseudo labels by directly predicting the event data. Simultaneously, we introduce offline event-to-image reconstruction into the framework, which generates another set of pseudo labels by predicting the reconstructed images. Through training on these hybrid pseudo labels, the network can progressively improve its ability to directly predict more accurate labels for event data.
To gradually mitigate the impact of low-quality reconstructed images during training, we approach this challenge as a noisy label learning (NLL) problem. In this context, we distinguish between noisy data (reconstructed images) and clean data (original events). Then, we introduce a noisy-label adaptation process to further refine pseudo labels at each iteration. In addition, due to the large domain gap between image and event, the network is prone to produce dispersed features in the target domain~\cite{gehrig2020video}. To counteract this issue, we also design a soft prototypical alignment (SPA) module to learn the intrinsic structure of the target domain and address the dispersion of target features.
%
%
As illustrated in Figure~\ref{fig:preview}, the proposed method is very effective, outperforming other state-of-the-art UDA approaches by a large margin and even surpassing several fully supervised methods. 

In summary, our contributions in this paper are:
\begin{itemize}
\item
We propose a hybrid pseudo-labeling framework for unsupervised event-based semantic segmentation. This framework gets rid of event-to-image pairs and is robust to noisy pseudo labels.
\item
We design a soft prototypical alignment (SPA) module to enforce the network to generate consistent event features under the same class, forming a more compact feature space in the target domain.
\item
Extensive experiments on two benchmark datasets demonstrate that our method outperforms previous state-of-the-art methods by a large margin. 
\end{itemize}

\section{Related Work}
\label{sec:Related Work}

\begin{figure*}[!t] 
\begin{center}
\includegraphics[width=0.9\linewidth]{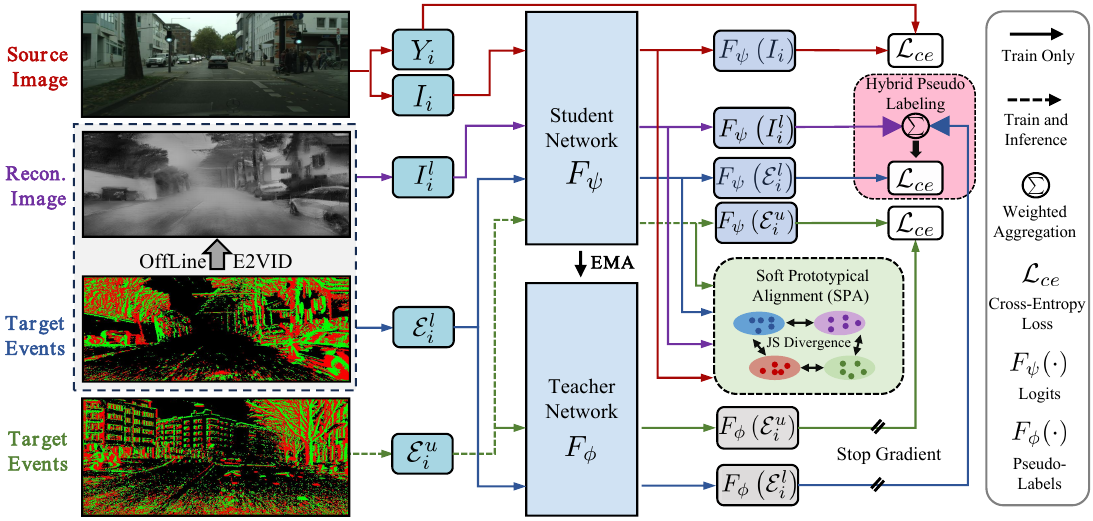}
\end{center}
\caption{Overview of the \modelname{} architecture. During training, we introduce offline event-to-image reconstruction as input to our framework. To avoid overfitting noise, we use only a small proportion (5\%) of the reconstructions. The network is trained by hybrid pseudo labels from reconstruction and self-prediction. Additionally, a soft prototypical alignment (SPA) module is designed to enhance the consistency of target domain features. In the inference phase, only events are used as input.}
\label{fig:overview}
\end{figure*}

\subsection{Event-based Semantic Segmentation}
Using deep learning, \cite{alonso2019ev} first introduces event cameras to the semantic segmentation task, with an architecture based on an encoder-decoder CNN, pre-trained on the well-known urban environment Cityscapes dataset~\cite{cordts2016cityscapes}. An open dataset, DDD17, containing annotated DAVIS driving records for this task is released in~\cite{binas2017ddd17}. 
%
\cite{gehrig2020video} enables the use of existing video datasets by transforming them into synthetic event data, facilitating the training of networks designed for real event data. Despite its capacity to leverage an unlimited number of video datasets, challenges persist due to the sim-to-real gap in many simulated scenarios. 
\cite{wang2021evdistill} employs two student networks for knowledge distillation from the image to the event domain. However, the method heavily depends on per-pixel paired events and active pixel sensor (APS) frames. Consequently, in scenarios where APS frames are unavailable, the application of such a knowledge distillation approach becomes significantly restricted.
\cite{wang2021dual} substitutes the active pixel sensor modality with grayscale images generated by E2VID~\cite{rebecq2019high}, transferring the segmentation task from the event domain to the image domain.
Recently, ESS~\cite{sun2022ess} addresses event-based semantic segmentation by introducing the DSEC-Semantic dataset, which relies on paired high-resolution images and events, thus providing high-quality semantic labels for event streams. ESS also introduces an event-to-image-based UDA method to transfer knowledge from the source image domain to the target event domain.

\subsection{Unsupervised Domain Adaptation}
Unsupervised domain adaptation (UDA) approaches can be divided into two key methodologies: domain adversarial learning and self-training. Domain adversarial learning focuses on aligning feature distributions across domains~\cite{ganin2016domain} but does not inherently ensure the discriminative power of target features~\cite{lv2019targan}.
In contrast, self-training capitalizes on a model’s high-confidence predictions to bolster the performance within the target domain. 
This approach significantly alleviates the domain shift issue by iteratively aligning the feature distribution of the target domain to match that of the source domain, which proves to be particularly effective in scenarios where obtaining labels for the target domain is challenging. In this context, strategies such as leveraging domain-invariant features~\cite{gong2013connecting, bui2021exploiting, li2020attribute}, pseudo-labeling~\cite{zhang2021prototypical, zou2018unsupervised}, intermediate domains~\cite{na2021fixbi, wang2022understanding,jing2023x4d}, and consistency regularisation~\cite{hou2016unsupervised} have been used.
We consider that under a similar task and scenario, event data can also be drawn close to RGB images semantically through the application of UDA methods.

\section{Method}
\label{sec:Method}
As illustrated in Figure~\ref{fig:overview}, the proposed \modelname{} framework incorporates self-training UDA techniques, described in Section~\ref{subsec:Framework}, and employs offline event-to-image reconstruction to generate hybrid pseudo labels, covered in Section~\ref{subsec:Labeling}.
To gradually mitigate the impact of low-quality and blurred areas in offline-reconstructed images, we introduce a noisy label learning (NLL) method to refine pseudo labels.
We further propose a soft prototypical alignment (SPA) module to explore the intrinsic structure of event data, alleviating the impact of feature divergence as detailed in Section~\ref{subsec:Consistency}.

\subsection{Definitions and Problem Formulation}
\label{subsec:Definitions}
%
In a UDA framework for event-based semantic segmentation, a neural network $F$ is usually trained from labeled source dataset $\mathcal{S}=\left\{{I}_i, {Y}_i\right\}_{i=1}^M$ to transfer to an unlabeled target dataset $\mathcal{T}=\left\{\mathcal{E}_i\right\}_{i=1}^N$. Specifically, the source domain $S$ consists of images $I_i \in \mathbb{R}^{H \times W}$ and their corresponding labels $Y_i \in \mathbb{R}^{H \times W}$. In contrast, the target domain $T$ consists of numerous continuous and asynchronous event streams $\mathcal{E}_i$ and without having access to the target labels $V_i$. Each event stream $\mathcal{E}_i$ can be represented as a series of tuples $\{(x_j,y_j,t_j,p_j)\}$, where $j$ denotes the sample index, $x$, and $y$ denote the spatial co-ordinates, $t$ represents the timestamp, and $p$ indicates the binary polarity (positive or negative) of brightness changes occurring between two timestamps.
Due to the high temporal resolution of $\mathcal{E}_i$, we sub-sample $\mathcal{E}_i$ into a sequence of voxel grid representations~\cite{zhu2019unsupervised}, where each voxel grid is constructed from non-overlapping temporal windows with a fixed number of events. These are then effectively superimposed to form a static frame.

\subsection{UDA Framework Overview}
\label{subsec:Framework}
We modify DaFormer~\cite{hoyer2022daformer} as the backbone and baseline for our event-based semantic segmentation UDA method. The framework is composed of two networks: a teacher network $F_\phi$ and a student network $F_\psi$. Other modules in DaFormer are eliminated to ensure the simplicity and efficiency of our method. 
To facilitate knowledge transfer from the source domain to the target domain, the modified baseline is trained using the mixed data of labeled images and unlabeled events.
To be specific, in our work, the student network $F_\psi$ first conducts warm-up by being trained with the supervised loss on the source image domain
\begin{equation}
{\mathcal{L}}_s(F_\psi \mid \mathcal{S})=\frac{1}{|\mathcal{S}|} \sum_{i=1}^{|\mathcal{S}|} H\left(F_\psi\left(I_i\right), Y_i\right),
\end{equation}
where $H$ denotes the cross entropy function.
Correspondingly, the parameters of the teacher network are updated using the exponential moving average (EMA)~\cite{tarvainen2017mean}  from the student model to maintain stability. After warm-up, the framework follows a self-training strategy, where the teacher network directly predicts the event data to generate pseudo labels for the training of the student model. This process is repeated until the networks have converged. 

In addition, augmentation methods, such as jitter and ClassMix~\cite{olsson2021classmix}, are used on both events and images to improve the method’s availability across domains. Although self-training UDA is usually an effective technique, it is challenging to obtain satisfactory results due to the large domain gap between images and events. Furthermore, it suffers from the aforementioned single-source noisy pseudo labels.

\subsection{Hybrid Pseudo-Labeling}
\label{subsec:Labeling}
%
To address the above issues, we consider the E2VID~\cite{E2VID} method to reconstruct the event streams into simulated images, which are then incorporated into our framework as an intermediate domain to narrow down the gap between the source image domain and the target event domain. The reconstructed images also provide another set of pseudo labels to alleviate the bias of single-source pseudo labels. In particular, we randomly sample the event dataset $\mathcal{T}=\left\{\mathcal{E}_i\right\}_{i=1}^N$ to create two groups, $\mathcal{T}_l=\left\{\mathcal{E}_i^l\right\}_{i=1}^a$ and $\mathcal{T}_u=\left\{\mathcal{E}_i^u\right\}_{i=1}^b$, where $a+b=N$. Event streams $\mathcal{E}_i^l$ are reconstructed into simulated images $I_i^{l}$ as
\begin{equation}
I_i^{l}=E2VID\left({\mathcal{E}}_i^l\right).
\end{equation} 
Now, the inputs to the student network encompass source images $I_i$, unlabeled events $\mathcal{E}_i^u$, unlabeled events $\mathcal{E}_i^l$ and the corresponding reconstructed images $I_i^{l}$. Notably, we do not reconstruct all event streams into simulated images to avoid the network overfitting these noisy data.

As illustrated in Figure~\ref{fig:overview}, the student network $F_\psi$ takes the reconstructed image $I_i^{l}$ as input and generates the predicted probability map. This map is then utilized as $F_\psi(I_i^l)$, the pseudo-ground-truth for $\mathcal E_i^l$.
%
%
%
Simultaneously, similar to the self-training backbone, event data $\mathcal{E}_i^u$ and $\mathcal{E}_i^l$ are input to the teacher network $F_\phi$ to obtain the direct pseudo labels $F_\phi(\mathcal{E}_i^u)$ and $F_\phi(\mathcal{E}_i^l)$. For $\mathcal{E}_i^u$, the student network $F_\psi$ is trained with the supervised loss ${\mathcal{L}}_u$ calculated as
\begin{equation}
{\mathcal{L}}_u(F_\psi \mid \mathcal{T}_u)=\frac{1}{|\mathcal{T}_u|} \sum_{i=1}^{|\mathcal{T}_u|} H\left(F_\psi\left(\mathcal{E}_i^u\right), F_\phi(\mathcal{E}_i^u)\right).
\end{equation}

For $\mathcal{E}_i^l$, $F_\phi(\mathcal{E}_i^l)$ together with the event pseudo-ground-truth $F_\psi(I_i^l)$ constitutes the hybrid pseudo labels.

\begin{figure}[!t] 
\begin{center}
\includegraphics[width=0.8\linewidth]{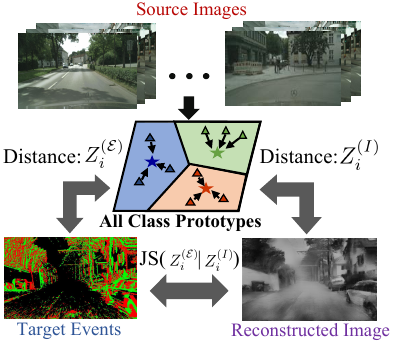}
\end{center}
\caption{The concept of our SPA module on source domain, reconstructed images, and events.}
\label{fig:SPA}
\end{figure}

The event-to-image reconstruction process suffers from limited interpretability and a lack of control, leading to low-quality reconstructed images $I_i^{l}$, \emph{e.g.}, incorrect content and blurred areas. Predicting semantic segmentation maps on these images and viewing them as pseudo labels will inevitably introduce significant noise. Directly using them during training may result in sub-optimal performance. 
Therefore, we treat this as a noisy label learning problem and explicitly regard $F_\psi(I_i^l)$ as a noisy label of the events $\mathcal{E}_i^l$. Inspired by~\cite{yu2023semi}, we employ a label correction strategy based on self-prediction to mitigate the noise issue. This strategy adapts the noisy distribution from the pseudo-ground-truth $F_\psi(I_i^l)$ to the view of the event distribution.
Specifically, for each $\mathcal{E}_i^l$, we reconstruct the refined pseudo label $\hat{V_i}^l$ by combining $F_\psi(I_i^l)$ and the $F_\phi(\mathcal{E}_i^l)$ as
\begin{equation}
\label{eq:nll}
\hat{V_i}^l=(1-\alpha) F_\psi(I_i^l)+\alpha F_\phi(\mathcal{E}_i^l) ,
\end{equation}
with ratio $\alpha$. Then, the modified loss ${L}_l$ for $\mathcal{E}_i^l$ is
\begin{equation}
{\mathcal{L}}_l(F_\psi \mid \mathcal{T}_l)=\frac{1}{|\mathcal{T}_l|} \sum_{i=1}^{|\mathcal{T}_l|} H\left(F_\psi\left(\mathcal{E}_i^l\right), \hat{V_i}^l\right).
\end{equation}
During training, the teacher network progressively generates more accurate $F_\phi(\mathcal{E}_i^l)$, gradually weakening the impact of $F_\psi(I_i^l)$.

\subsection{Soft Prototypical Alignment}
\label{subsec:Consistency}
By employing man-made paired $\mathcal{E}_i^l$ and $I_i^l$ to bridge the image domain and event domain, we aim to enhance the alignment between the source and target. However, using $F_\psi(I_i^l)$ as a pseudo-label may still not be able to solve the distribution misalignment because of the obvious differences in the distributions of $I_i^l$ and $\mathcal{E}_i^l$.
%
%
Inspired by~\cite{zhang2021prototypical}, we propose a soft prototypical alignment (SPA) module to explicitly align the distributions for our problem. As illustrated in Figure~\ref{fig:SPA}, we employ the mean value $F_\psi(I_i)$ of each class on source images as prototypes $\eta$ and aim to align the prototype distance between $F_\psi(I_i^l)$ and $F_\psi(\mathcal{E}_i^l)$. 
The distance between $F_\psi(I_i^l)$ and $\eta$ is calculated as
\begin{equation}
Z_i^{(I)}=\frac{\exp \left(-\left\|{{F_\psi}\left(I_i^l\right)}-\eta \right\| / \tau\right)}{\sum \exp \left(-\left\|{{F_\psi}\left(I_i^l\right)}-\eta \right\| / \tau\right)},
\label{Zu}
\end{equation}
where $\tau$ is the coefficient temperature.
Similarly, the distance between $F_\psi(\mathcal{E}_i^l)$ and $\eta$ is calculated as
\begin{equation}    
Z_i^{(\mathcal{E})}=\frac{\exp \left(-\left\|{{F_\psi}\left(\mathcal{E}_i^l\right)}-\eta \right\| / \tau\right)}{\sum \exp \left(-\left\|{{F_\psi}\left(\mathcal{E}_i^l\right)}-\eta \right\| / \tau\right)}.
\label{Zi}
\end{equation}
We use the Jensen-Shannon (JS) divergence~\cite{englesson2021generalized} instead of KL divergence used in ~\cite{pan2019transferrable} for distribution alignment due to the symmetry of JS divergence. This ensures an equal pulling effect on the distributions of $F_\psi(I_i^l)$ and $F_\psi(\mathcal{E}_i^l)$. The JS divergence loss is calculated as
\begin{equation}
{\mathcal{L}}^S_{JS}=\operatorname{JS}\left(Z_i^{(I)} \| Z_i^{(\mathcal{E})}\right),
\label{JL}
\end{equation}
and compels the network to generate consistent event features and image features for $\mathcal{E}_i^l$ and $I_i^l$ under the same class.

Additionally, $\mathcal{E}_i^l$ and $\mathcal{E}_i^u$ are not trained with the same set of pseudo labels, making the target distributions of $\mathcal{E}_i^l$ and $\mathcal{E}_i^u$ more likely to be dispersed. In such a scenario, the network fails to rectify the labels of target data located at the far end of the class cluster. Considering that the distributions of $F_\psi(\mathcal{E}_i^l)$ and $F_\psi(\mathcal{E}_i^u)$ belong to the same scene, their distributions are expected to exhibit similar relative distances. To achieve this, we further employ the mean value of each class in $F_\psi(I_i^l)$ as a prototype. We then bring in the relative distances of $F_\psi(\mathcal{E}_i^l)$ and $F_\psi(\mathcal{E}_i^u)$ to $F_\psi(I_i^l)$, respectively. Employing a methodology akin to Eqns.~(\ref{Zu}), (\ref{Zi}), and~(\ref{JL}), we obtain ${\mathcal{L}}^I_{JS}$ that forms a more compact feature space in the target domain.

The overall loss in our framework is defined as
\begin{align}
\mathcal{L}= \mathcal{L}_s + \mathcal{L}_u + \mathcal{L}_l + \omega ({\mathcal{L}}^S_{JS} + {\mathcal{L}}^I_{JS}),
\label{eq:loss_final}
\end{align}
where $\omega$ denotes a hyper-parameter.

\section{Experiments}
\label{sec:Experiment}

\subsection{Dataset}
\label{sec:Data}
As target data, we evaluate the proposed framework on two event-based semantic segmentation datasets, namely DSEC-Semantic~\cite{gehrig2021dsec} and DDD17~\cite{binas2017ddd17}. These driving-focussed datasets were captured using automotive-grade event cameras, encompassing a diverse range of urban and rural settings.

The DDD17 dataset comprises per-pixel paired events and frames captured by DAVIS event cameras with a resolution of $346 \times 260$. In~\cite{alonso2019ev}, semantic labels were generated using pre-trained segmentation networks based on DAVIS images, resulting in 15,950 samples for training and 3,890 for testing. Due to the low resolution, several categories in DDD17 have been merged into six classes, namely flat (road and pavement), background (construction and sky), object, vegetation, human, and vehicle. 

DSEC-Semantic, a recently introduced dataset for event-based semantic segmentation, extends the comprehensive DSEC dataset~\cite{gehrig2021dsec}. It includes 53 driving sequences captured by an event camera at a resolution of $640 \times 480$. \cite{sun2022ess} used a state-of-the-art image-based segmentation method~\cite{tao2020hierarchical} to generate segmentation labels. This process yields 8,082 labeled training samples and 2,809 testing samples, distributed across 11 classes: sky, building, fence, person, road, pole, sidewalk, vegetation, vehicle, wall, and traffic sign.

As source data, we use the CityScapes street scene dataset~\cite{cordts2016cityscapes}, which includes 2,975 training and 500 validation images with a resolution of $2048 \times 1024$. Following common practice in UDA methods, we resize the CityScape images to $1024 \times 512$ pixels.

\begin{table*}[htbp]
\centering
\caption{Performance and necessary number of events on DSEC-Semantic dataset in both UDA and fully supervised learning settings.}
\label{tab:DSEC}
\begin{tabular}{c|c|c|ccc}
\hline
Type       & Method       & No. of Events & Accuracy {[}\%{]} & mIoU {[}\%{]} \\ \hline
\multicolumn{1}{l|}{Supervised} & \multicolumn{1}{l|}{EV-SegNet~\cite{alonso2019ev}}               &    -        &                   88.61             & 51.76                \\
           & \multicolumn{1}{l|}{HALISE~\cite{biswas2022halsie}}               &   -         &                   89.01             & 52.43               \\
           & \multicolumn{1}{l|}{ESS~\cite{sun2022ess}}                     &    2E6        &                   89.37             & 53.29                \\ \hline
\multicolumn{1}{l|}{UDA}        & \multicolumn{1}{l|}{EV-Transfer~\cite{messikommer2022bridging}}        & 2E6        & 60.50              & 23.20        \\
           & \multicolumn{1}{l|}{E2VID~\cite{rebecq2019high}}       & 2E6        &  76.67       & 40.70  \\
           & \multicolumn{1}{l|}{ESS~\cite{sun2022ess}}           & 2E6          &  84.04       & 44.87                                 \\ 
        & \multicolumn{1}{l|}{{\bf Ours}}                &   {\bf 1.8E5 ($\downarrow$ 91.0\%)}        &                   {\bf 89.92 (+ 5.88\%) }          & {\bf 55.19 (+10.32\%) }              \\ \hline
\end{tabular}
\end{table*}

\subsection{Implementation Details}
\label{sec:detail}
In our experiments, we employ DaFormer~\cite{hoyer2022daformer} as our UDA backbone. The encoder in Daformer uses an MiT-B5 model~\cite{xie2021segformer} and is pre-trained on ImageNet-1K. Across all experiments, the batch size is consistently set to 4. We use the AdamW optimizer with a weight decay of $1\times 10^{-4}$. The learning rate is set to $6\times 10^{-5}$ and we use a learning rate warm-up for 1,500 iterations, with a linear increase in the learning rate during this period. We additionally warm-up for 5,000 iterations on the source dataset to make the network gain the initial semantic segmentation ability.
$\alpha$ in Eq.~\ref{eq:nll} and $\omega$ in Eq.~\ref{eq:loss_final} are both set to 0.5. For data augmentation in both source and target domains, we follow~\cite{XieLLLHW23,hoyer2022daformer} and employ techniques such as color jitter, Gaussian blur, and ClassMix~\cite{olsson2021classmix}. These augmentations are instrumental in training the model to learn more robust features across different domains.

In the event-to-image simulation process, Spade E2VID~\cite{rebecq2019high} is employed as our emulator for reconstruction. This step occurs solely in the offline phase, ensuring that it does not impact the efficiency of our online training and testing process. It is worth noting that E2VID will progressively produce expanding black artifacts if fed with discontinuous event inputs. To mitigate this problem, we reinitialize the E2VID network each time an image is reconstructed, preventing the occurrence of such artifacts. Regarding the event pre-processing on the DDD17 dataset, events are converted into 20 voxel grids, with each grid containing 32,000 events. For the DSEC-Semantic dataset, due to its higher resolution, the number of voxel grids is increased to 40, and each grid comprises 100,000 events.

\subsection{Comparison with State-of-the-Art}
\label{sec:Comparison}
We compare our method with previous relevant approaches, and use the top-1 accuracy and the mean intersection over union (mIoU) as the common semantic segmentation evaluation metrics. 
Beyond UDA methods, certain approaches have embraced a fully supervised setting to tackle the challenges. EV-SegNet~\cite{alonso2019ev} presents the first baseline for event-based semantic segmentation, which employs an encoder-decoder architecture and takes only events for fully supervised learning. HALISE~\cite{biswas2022halsie} encodes event frames and source images into a spike stream, representing information in a binarised manner, and aligns the feature distribution in these spike streams. EV-Transfer~\cite{messikommer2022bridging} fabricates the motion of a still image to generate event streams, and then uses source labels and the corresponding synthetic events to conduct training. E2VID~\cite{rebecq2019high} converts events in the DSEC-Semantic dataset to reconstruct images, and then predicts semantic segmentation maps using other pre-trained models. E2VID can only perform direct transfer as there is no event label for training. VID2E~\cite{gehrig2020video} converts source video frames to synthetic events and trains on the source labels. ESS~\cite{sun2022ess} employs the above E2VID-based process to generate pseudo-labels and attempts to transfer knowledge from the source image domain to the target event domain by the UDA technique.
While methods employing supervised learning may achieve superior results compared to traditional UDA approaches, their reliance on labels significantly elevates the demands for dataset collection.

{\bf DSEC-Semantic dataset}. 
We employ the CityScape dataset as the labeled source dataset and the DSEC-Semantic dataset as the unlabeled target dataset. This dataset poses additional challenges due to its more fine-grained categories compared to the DDD17 dataset. We report the obtained results for all methods in~Table~\ref{tab:DSEC}.

\begin{figure}[!b] 
\begin{center}
\includegraphics[width=1.0\linewidth]{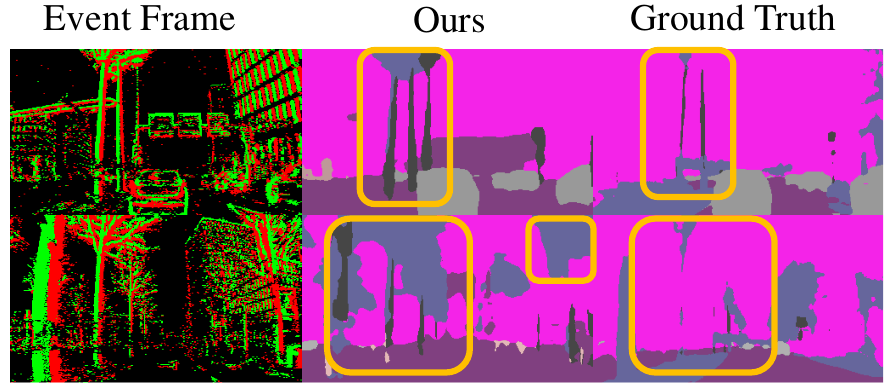}
\end{center}
\caption{Example results on DDD17 dataset. The DDD17 ground truth lacks details for some objects.}
\label{fig:ddd17}
\end{figure}

As we can see from there, our method demonstrates a significant improvement, outperforming the previous state-of-the-art UDA work ESS by 5.87\% and 9.65\% in terms of accuracy and mIoU, respectively. Notably, our UDA-based method even surpasses the performance of fully supervised approaches by 0.55\% in terms of accuracy and 1.9\% in terms of mIoU. Since this is a highly imbalanced dataset, the gain in mIoU is more representative in the segmentation task. In addition, by solely utilizing the E2VID reconstruction method offline, our approach avoids dependency on recurrent networks in E2VID during both training and inference, significantly reducing the required input events from 2E6 to 1E5 (a 95\% reduction). These enhancements remarkably prove the effectiveness and computational efficiency of our proposed method.

\begin{figure*}[!t] 
\begin{center}
\includegraphics[width=1\linewidth]{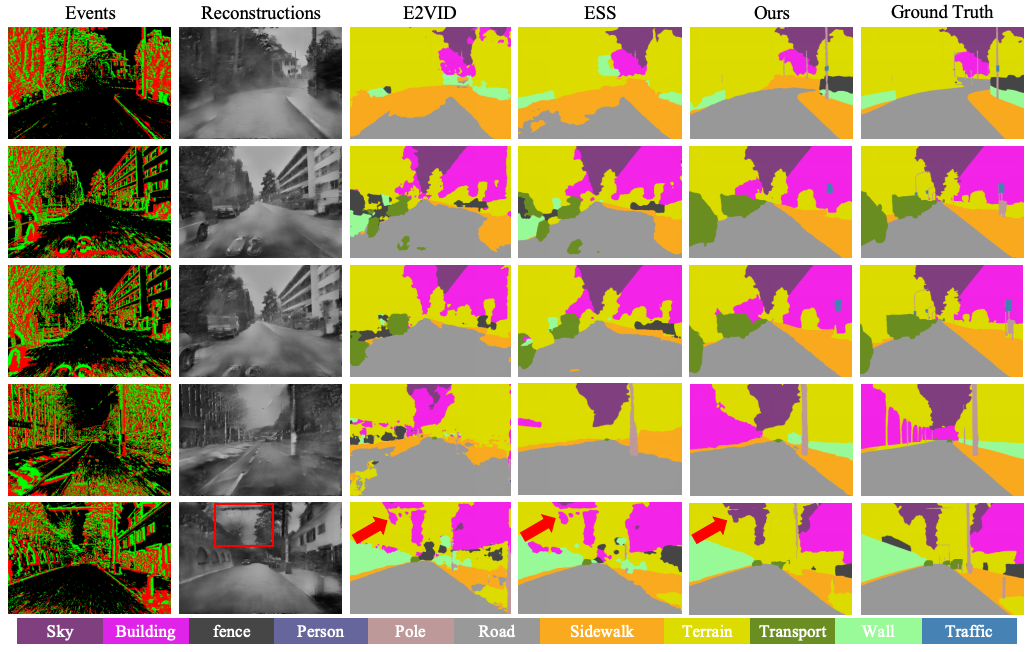}
\end{center}
\caption{Visualization results on DESC-Semantic dataset. From left to right: event frame, event-to-image reconstruction, the maps predicted by E2VID, ESS, and our proposed \modelname{}, ground truth.}
\label{fig:visual}
\end{figure*}

Some example results are visualized in Figure~\ref{fig:visual}. The background of the reconstructed image exhibits fuzzy regions and low resolution, which inevitably poses significant challenges to semantic segmentation Networks. For instance, due to the lack of texture information in events, the reconstructed sky category appears very similar to the building category in terms of contrast and edge information, leading to potential misinterpretation of the model's predictions (as indicated by the red arrow). The proposed hybrid pseudo-labeling method effectively mitigates these interference factors in reconstructed images, resulting in improved performance.

{\bf DDD17 dataset}. 
Table~\ref{tab:tab1} reports the UDA results on the DDD17 dataset for event-based semantic segmentation. Similar to the DSEC-Semantic dataset, only labeled images from CityScape and unlabeled events from DDD17 are available in this task. Table~\ref{tab:tab1} showcases that our method achieves consistent optimal results, outperforming the previous state-of-the-art work by 1.05\% (mIoU) and 0.79\% (accuracy), respectively.

Since the event ground truth in DDD17 is derived from the low-quality paired images, this significantly impacts the reliability, especially concerning texture details, as also mentioned in~\cite{sun2022ess}. As Figure~\ref{fig:ddd17} illustrates, our predictions even surpass the ground truth in object details. Taking the first line in Figure~\ref{fig:ddd17} as an example, in the yellow box, our network better separates the details of the trees and streetlights, which are missing in the DDD17 ground truth. Similarly, in the second line, our method segments more correct trees, while the DDD17 ground truth misclassifies trees with sky. This discrepancy could potentially lower our performance during evaluation. Due to the higher resolution and quality of the DSEC-Semantic dataset, we opted for this dataset to evaluate our method and comparison works.

\begin{table}[]
\centering
\caption{Performance comparison of \modelname, with state-of-the-art methods on DDD17 dataset in UDA setting. Only source labels are available. }
\label{tab:tab1}
\begin{tabular}{ccc}
\hline
Method      & Accuracy {[}\%{]} & mIoU {[}\%{]} \\ \hline
\multicolumn{1}{l|}{EV-Transfer~\cite{messikommer2022bridging}}  & 47.37         & 14.91     \\
\multicolumn{1}{l|}{E2VID~\cite{rebecq2019high}}      & 83.24         & 44.77     \\
\multicolumn{1}{l|}{VID2E~\cite{gehrig2020video}}       & 85.93             & 45.48         \\
\multicolumn{1}{l|}{ESS~\cite{sun2022ess}}         & 87.86        & 52.46     \\ \hline
\multicolumn{1}{l|}{{\bf Ours}}        &    {\bf 88.65 (+0.79\%)}          &     {\bf 53.51+(1.05\%)}          \\ \hline
\end{tabular}
\end{table}

\subsection{Comprehensive Analysis}
\label{sec:Ablation}
Since DSEC-Semantic is a higher-quality dataset, all ablation experiments are conducted on DSEC-Semantic.

{\bf Design analysis of our framework.}
We conduct several ablation studies to assess the effectiveness of the proposed framework. As depicted in Table~\ref{tab:ablation_main}, (a) directly applying the UDA baseline alone does not yield satisfactory results, likely due to the substantial domain gap between the image and event domains.
Similarly, (b) training directly on the event-to-image (ETI) reconstructed images from E2VID also results in unsatisfactory performance. This result verifies the aforementioned discussion, namely event-to-image-based methods will suffer from the noise brought by the reconstructed image. Both (a) and (b) highlight the unreliability of solely relying on the single-source pseudo labels and emphasize the necessity for hybrid label learning.
An intriguing observation is that (c) employing the source data to pre-train the network for a certain number of iterations, \emph{i.e.}, using a warmup phase, significantly enhances the performance.
In (d), our method is based on source domain warm-up, and as described in Section~\ref{sec:Method}, the E2VID reconstructed images are introduced on top of the UDA backbone to provide the hybrid pseudo labels for events, leading to considerable performance gains.
\begin{table}[htbp]
\centering
\caption{Ablations study on DSEC-Semantic dataset.}
\label{tab:ablation_main}
\resizebox{\columnwidth}{!}{%
\begin{tabular}{c|ccccc|c}
\hline
Method & Baseline & ETI & Warmup & NLL & SPA & mIoU {[}\%{]} \\ \hline
(a)    & $\checkmark$    &       &        &     &     & 36.76         \\
(b)    &     & $\checkmark$      &        &     &     & 40.70         \\
(c)    & $\checkmark$    &       &   $\checkmark$     &     &     & 44.87         \\
(d)    & $\checkmark$    & $\checkmark$      & $\checkmark$       &     &     & 51.08          \\ \hline
(e)    & $\checkmark$    & $\checkmark$      & $\checkmark$       & $\checkmark$    &     & 52.23            \\
(f)    & $\checkmark$    & $\checkmark$      & $\checkmark$       &    & $\checkmark$    & 52.69            \\ 
{\modelname, }   & $\checkmark$    & $\checkmark$      & $\checkmark$       & $\checkmark$    & $\checkmark$    & {\bf 55.19 }           \\ \hline
\end{tabular}
}
\end{table}

\begin{table}[!t]
\centering
\caption{Ablation study for the proportion of event samples participating in the event-to-image reconstruction.}
\label{tab:ablation_pro}
\begin{tabular}{c|cc}
\hline
Proportion & Accuracy {[}\%{]} & mIoU {[}\%{]} \\ \hline
0\%        & 82.71             & 44.87         \\
1\%        & 86.54             & 46.84         \\ \hline
{\bf 5\%}        & {\bf 89.91}             & \textbf{55.19}         \\
10\%       & 89.89             & 55.15         \\
50\%       & 89.81             & 54.96         \\
80\%       & 89.75             & 54.82         \\
100\%      & 89.63             & 54.51         \\ \hline
\end{tabular}
\end{table}

We further validate the effectiveness of the proposed NLL strategy and SPA module. As shown in Table~\ref{tab:ablation_main}, NLL reduces the noise of pseudo-labels on reconstructed images through iterative label refinement, making it more adaptive to the event domain and resulting in a performance gain. 
SPA prioritizes the divergence of various features on the target domain and aligns the labeled and unlabeled events with the source domain prototype, contributing to enhanced evaluation performance. Ultimately, the simultaneous introduction of these two modules in our framework leads to optimal performance.

{\bf Proportion of reconstructed event samples.}
In our framework, we do not transform all event data into reconstructed images to avoid overfitting the reconstruction noise. In fact, as demonstrated in Table~\ref{tab:ablation_pro}, the optimal result is achieved when using only 5\% of the event data to generate the reconstructed images as the pseudo labels. Performance experiences a slight decline as more reconstructed images are introduced. Particularly, when using 100\% of the data, it results in an mIoU drop to 54.51\%. The lower dependence on the number of reconstructed images also underscores the remarkable computational efficiency of our method during training. Further reduction of the ratio, below 5\%, leads to progressively worse performance, reaching its lowest point at a 0\% ratio and reverting back to the UDA backbone.


{\bf Online/Offline reconstruction pseudo label.}
Offline event-to-image reconstruction enables us to directly predict the reconstructed image using a pre-trained network and get pseudo labels for event data, which are named reconstruction pseudo labels here. In this section, we compare the effects of fixing these reconstruction pseudo labels and iteratively repredicting them by our network during training. As shown in Table~\ref{tab:Online}, the online reprediction strategy remarkably surpasses the offline fixing strategy, demonstrating that our method becomes more powerful during training and can predict more accurate reconstruction pseudo labels for event data.  

\begin{table}
\centering
\caption{Ablation study for online and offline reconstruction pseudo labels.}
\label{tab:Online}
\begin{tabular}{l|cc}
\hline
\multicolumn{1}{c|}{Method} & Accuracy {[}\%{]} & mIoU {[}\%{]} \\ \hline
OffLine                     & 83.25             & 48.45         \\
{\bf OnLine }                     & {\bf 89.92}             & {\bf 55.19}         \\ \hline
\end{tabular}
\end{table}

\section{Discussion and Limitation}

Due to the imbalance issue presented in the benchmark datasets, the accuracy performance of classes with insufficient samples, \emph{e.g.}, 'rider' and 'traffic light,' is comparatively lower than the accuracy of some other classes, \emph{e.g.}, sky and road. These results are illustrated in the visualization examples in {\bf Supplementary Materials}. Despite that our approach yields significant improvement for the classes with a small number of samples when compared to previous methods, we will further consider more strategies to deal with the data imbalance issue in the future work.




\section{Conclusion}
In this paper, we have proposed a novel hybrid pseudo-labeling framework \modelname{} for unsupervised event-based semantic segmentation. \modelname{} effectively alleviates the challenges posed by noisy pseudo labels, a common issue in this field. The proposed method uniquely incorporates self-training unsupervised domain adaptation and offline event-to-image reconstruction to generate high-quality hybrid pseudo labels. The introduction of a noisy label learning strategy further refines the pseudo labels gradually. Moreover, a soft prototypical alignment (SPA) module significantly enhances the consistency and reliability of the target features. The effectiveness of \modelname{} is evidenced by its superior performance in extensive experiments, where it not only surpasses existing state-of-the-art UDA methods but also exceeds several supervised methods. 

\section{Acknowledgments}
This work is supported by the Shanghai AI Laboratory, National Key R\&D Program of China (2022ZD0160101), the National Natural Science Foundation of China (62376222), and Young Elite Scientists Sponsorship Program by CAST (2023QNRC001). This work is supported by 111 Project (No. D23006).

{
    \small
    \bibliographystyle{ieeenat_fullname}
    \bibliography{main}
}


\end{document}